\documentclass[runningheads]{llncs}
\usepackage[T1]{fontenc}
\usepackage{caption}
\usepackage{subfigure}
\usepackage{cite}
\usepackage{amsmath,amssymb,amsfonts}
\usepackage{algorithmic}
\usepackage{graphicx}
\usepackage{textcomp}
\usepackage{xcolor}
\usepackage{multirow}
\usepackage{booktabs}
\usepackage{wrapfig}
\usepackage{marvosym}

\usepackage{hyperref}

\begin{document}
	\title{Features Fusion Framework for Multimodal Irregular Time-series Events}
	
	%
	
	\author{Peiwang Tang\inst{1, 2} \and
	Xianchao Zhang \inst{3, 4(}\textsuperscript{\Letter}\inst{)}
	}
	%
	%
	\institute{Institute of Advanced Technology, University of Science and Technology of China, Hefei 230026, China\\
			\email{tpw@mail.ustc.edu.cn} \and
		G60 STI Valley Industry \& Innovation Institute, Jiaxing University, Jiaxing 314001, China \and
		Key Laboratory of Medical Electronics and Digital Health of Zhejiang Province, Jiaxing University, Jiaxing 314001, China\\
		\email{zhangxianchao@zjxu.edu.cn} \and
		Engineering Research Center of Intelligent Human Health Situation Awareness of Zhejiang Province, Jiaxing University, Jiaxing 314001, China
	}
	\maketitle              
\begin{abstract}
Some data from multiple sources can be modeled as multimodal time-series events which have different sampling frequencies, data compositions, temporal relations and characteristics. Different types of events have complex nonlinear relationships, and the time of each event is irregular.
Neither the classical Recurrent Neural Network (RNN) model nor the current state-of-the-art Transformer model can deal with these features well.
In this paper, a features fusion framework for multimodal irregular time-series events is proposed based on the Long Short-Term Memory networks (LSTM). 
Firstly, the complex features are extracted according to the irregular patterns of different events. Secondly, the nonlinear correlation and complex temporal dependencies relationship between complex features are captured and fused into a tensor. Finally, a feature gate are used to control the access frequency of different tensors.	
Extensive experiments on MIMIC-III dataset demonstrate that the proposed framework significantly outperforms to the existing methods in terms of AUC (the area under Receiver Operating Characteristic curve) and AP (Average Precision).

\end{abstract}
\section{Introduction}
In general terms, a modality refers to the way in which something happens or is experienced \cite{baltruvsaitis2018multimodal}.
To our best knowledge, many existing works have demonstrated that Neural Network can achieve an excellent result in single modality processing such as image classification \cite{tan2019efficientnet}, speech synthesis \cite{liu2021vara}, natural language processing \cite{vaswani2017attention}. 
In the field of data, multimodal is used to represent different forms of data, or different formats of the same form, which generally represents text, picture, audio and video \cite{liu2020urban, thistwo}. 
Hence, multimodal data processing have attracted a wide attention from the academia, especially for multimodal fusion which is one of the original topics in multimodal machine learning \cite{baltruvsaitis2018multimodal}. Neural Networks is expected to tackle the multimodal fusion problem  \cite{ngiam2011multimodal} and has been used extensively to fuse  information for text, image and audio \cite{liu2018efficient, nagrani2021attention}, gesture recognition \cite{neverova2015moddrop}, and video or image description generation \cite{ramesh2021zero, wu2021n}, since the earliest investigation of AVSR \cite{potamianos2003recent}.
However, almost of all these studies focus on text, images or speech modes rather than multimodal time-series which is a critical ingredient across many domains, so how to effectively process multimodal data still need further study. 
Many methods have been launched to process simple single mode time-series data \cite{armandpour2021deep, zhou2021informer}, which have achieved the best result in their respective field. But they have no way to directly use multimodal time-series data, for example multisensor data, medical time-series data.

The problem of features fusion is challenging in multimodal irregular time-series data processing\cite{fu2008multiple}. For example, for clinical data, patient’s electronic health records can be abstracted into thousands of interrelated medical events with temporal information, including complex allergy history, family genetic history, drug list, hospitalization records and other historical records. Different event has almost absolutely different frequency of recording. E.g. patient’s hospitalization records may be only once a few years, but medication records could be many times a year. Not only different events have different recording frequencies, but also the same type events have significant differences in their different nature. For example, attributes of drug taking events such as drug type, dose and test events include specific indicators and comparison results with normal range values. In order to integrate the features of these events, we must describe these dependencies.

In order to solve the above problems in multimodal irregular time-series events, in this paper, the following contributions is presented in this paper:
(1) We propose a new features fusion method to deal with multimodal data, where the features of complex data are fused into a common feature subspace. This method can be applied to different multimodal data.
(2) We explore different encoding methods for temporary features, and found a method to embed the temporary features into the non-temporary features, which allows us to better deal with time-series data
(3) We propose a model called FG-LSTM which developed from the Recurrent Neural Network such as Phased LSTM \cite{neil2016phased} to deal with the problem of irregular time-series data. Our proposed model filters the input features by feature gate while recording the complex temporal relationship between different features.
(4) We compare with other models, and the experiment results based on the real data demonstrate that the prediction performance of our model is significantly improved.

\section{Related Work }

\subsection{Multimodal Fusion problem}
Multimodal fusion mainly refers to the comprehensive processing of multimodal data by computer, which is responsible for fusing the information of each mode to perform target prediction \cite{thistwo}.
Tensor Fusion Network (TFN) \cite{zadeh2017tensor}is a multimodal network for features fusion through matrix operation to directly fuse the three features vectors of the data with three modes (such as text, image and audio). However, since TFN calculates the correlation between the elements of different modes through the tensor outer product between modes, it will greatly increase the dimension of features tensor and result in a too large model that is difficult to train.
Low-rank Multimodal Fusion \cite{liu2018efficient} uses a low rank matrix to decompose the weight, and hence the TFN process is changed into a single linear transformation of each mode. Then the received multi-dimensional point by Low-rank Multimodal Fusion can be regarded as the sum of multiple low rank vectors, and thus the number of parameters in the model is reduced. Although Low-rank Multimodal Fusion is an upgrade of TFN, once the features are too long, it is still easy to explode parameters.
Multimodal Adversarial Representation Network \cite{li2020adversarial} adds a dual discriminator countermeasure network based on multimodal fusion (ordinary attention fusion), which captures dynamic commonness and invariance respectively.
Multimodal Bottleneck Transformer \cite{nagrani2021attention} uses a shared token between two Transformer, so that this token becomes a communication bottleneck of different modes to save computational attention. In this way, multimodal interaction can be limited to several shared tokens.
Compared with the above researches, we pay more attention to multimodal time-series events, and the above researches can also be regarded as special cases of multimodal time-series events.
\subsection{Time-series Forecasting}
Recurrent neural network (RNN) is a neural network used to process sequence data. 
Theoretically, RNN can store long-term memory and update the previous state according to the current input at any time, but in fact, it is very difficult. In another word, RNN is difficult to solve the problem of long-term dependence \cite{hochreiter2001gradient}.
LSTM \cite{hochreiter1997long} is a special RNN, which is mainly to solve the problems of gradient disappearance and gradient explosion in the process of long sequence training. Compared with ordinary RNN, LSTM has better performance in long sequences, but LSTM can only maintain a long-term dependence within about 50 time steps. 
Phased LSTM \cite{neil2016phased} can solve the problem that LSTM can not process irregular input sequences. By integrating different sampling frequencies or irregularly sampled data on phase gate, Phase LSTM can remember signals with different periods, and the state can propagate for a long time. When the processing sequence reaches thousands of steps, LSTM is almost unavailable, while Phased LSTM performs well. But Phased LSTM is not suitable for modeling the complex event sequence with thousands of event types. 
HE-LSTM \cite{LuchenLIU2019} is proposed to deal with heterogeneous temporal events in long-term dependence, but it can only extract event types while the features relationship of events can not be obtained. 
Transformer \cite{vaswani2017attention} is a powerful architecture that can achieve excellent performance on a variety of sequential learning tasks, which does not perform recursion on the sequence, but processes the feedforward model of the whole sequence simultaneously.
Recent research shows that transformer has the potential to improve the prediction ability \cite{tsai2019transformer}. However, transformer has some serious problems that make it unable to be directly applied to multimodal irregular time-series data, such as quadratic time complexity, high memory utilization and the inherent limitations of encoder-decoder architecture \cite{zhou2021informer}.
In addition to the above problems, the biggest problem of Transformer is that the model contains no recurrence and no convolution, which results in the input tensor can not contain the time relationship of the input sequence effectively\cite{dai2019transformer}.

\section{Methodology}

\begin{figure*}[t]
	\centering
	\includegraphics{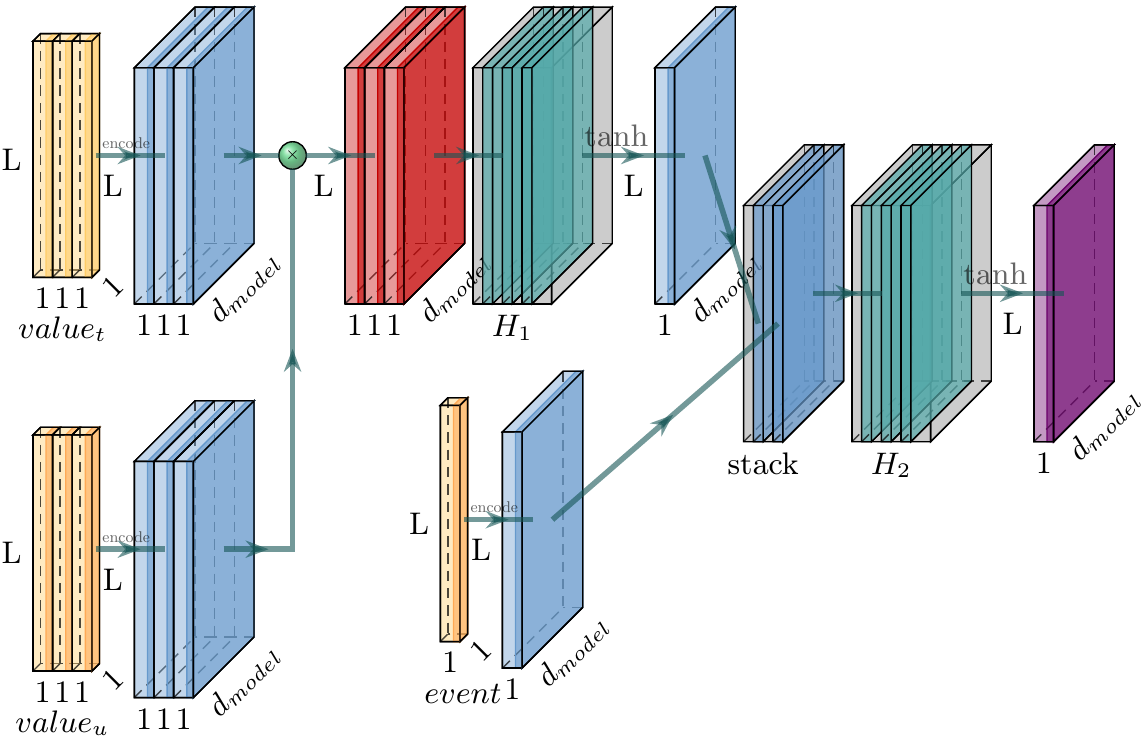}
	\caption{The non-temporal features fusion method.}
	\label{img1}
\end{figure*}

For time-series $\textbf{S}$ in a given scene, the features of the sequence are consist of dynamic events $\left \{ \textbf{E}_{t} \right \}_{1\leq t\leq L}$ with length $L$, and each event occurs at the same or different time. We arrange it according to the chronological order of events. Among the events that occur at the same time, the events recorded earlier are arranged in front.
Each time-series \textbf{S} corresponds to a discrete label, which indicates the state of the object at a certain time in the future. For example, in Clinical Endpoint Prediction Task, 0 or 1 indicates the patient's status (death or not) at a certain point in time in the future.
So prediction of the results of time-series \textbf{S} with this corresponding discrete label is defined as the classification of time-series \textbf{S}.

\subsection{Features fusion}

	Each time-series \textbf{S} contains many types of events, and each event has its own time of occurrence. We use $\mathcal{S}$ to represent the feature space where \textbf{S} is located, $\mathcal{E}$ to represent non-temporal information, that is, the feature space where the events is located, and $\mathcal{T}$ to represent the feature space where temporal information is located.
Formally, a sequence $ S = \left [ E_{1} , E_{2}, \ldots , E_{L}\right ] $, defines each element as
$E_{i} = \left ( e_i, t_i \right ) $, with $ e_i \in  \mathcal{E} $ being the non-temporal features at time $i$ and $t_i \in \mathcal{T} $ as an temporal features, and $t_i$ is the interval between the occurrence time of this event and the time when the first event of this time-series occurs. 
The features vector are defined over a joint space :
$\mathcal{S} := \left ( \mathcal{E} \times \mathcal{T} \right ) $.
The resulting permutation-invariant set is: 
$ \textbf{S}_{E} = \left\{ E_{1}, E_{2}, \ldots , E_{L}\right\} = \left\{ \left ( e_1, t_1 \right ) , \left ( e_2, t_2 \right ), \ldots ,\left ( e_L, t_L \right )\right\} $.
For each event we define $ e_i = \left ( type, attribute \right )$, where $type$ is the type of event, we use $\mathcal{F}_t$ to represent the feature space where type is located;  $attribute$ is the attribute of the event, we use $\mathcal{F}_a$ to represent the feature space where attribute is located. 
So the feature space of event :
$\mathcal{E} := \left ( \mathcal{F}_t \times \mathcal{F}_a \right ) $,
$\mathcal{E}$ is obviously a joint space, where $\mathcal{F}_t$ is the discrete feature space and $\mathcal{F}_a$ is the continuous feature space.
Similarly, attribute consists of two parts: $attribute = \left (value_t, value_u\right )$, where $value_t$ is the type of attribute, and $value_u$ is the specific value of attribute. The feature vector of attribute are defined over a joint space :
$\mathcal{F}_a := \left ( \mathcal{V}_t \times \mathcal{V}_u \right ) $,
where $\mathcal{V}_t$ is the discrete feature space of $value_t$ and $\mathcal{V}_u$ is the continuous feature space of $value_u$.

For each type of event, it can contain multiple types of attributes. While for different types of events, it may contain the same type of attributes or different types of attributes.
Therefore, it is difficult to find the feature space of events directly, and we need to characterize the complex relationship between different events. We demonstrate the non-temporal features fusion method as shown in Fig.~\ref{img1}, where $d_{model}$ is the encoded dimension:
(1) Select the first three-dimensional feature of attribute, fill up the deficiencies with 0, encode $value_t$ and $value_u$ as $V_t$, $V_u$ respectively, and then use $V_t \times V_u$ to get a new three-dimensional feature; 
(2) Use $1 \times 1$ convolution kernel to increase the dimension of the features obtained in the previous step, and then use $1 \times 1$ convolution kernel to reduce the dimension to one-dimensional features after being processed by the $tanh$ activation function;  
(3) Stack the features obtained in the previous step with the features encoded by event, then use $1 \times 1$ convolution kernel to increase the dimension, after processing by the $tanh$ activation function, use $1 \times 1$ convolution kernel to reduce the dimension to obtain the one-dimensional non-temporal features.
For $\mathcal{V}_u$ of continuous feature space, we do not simply encode $value_u$ with convolution or fully connected layers, instead encode $value_u$ with the help of $\mathcal{V}_t$ of discrete feature space, as shown in the formula:
\begin{equation}V_u =W_{v_t}  \times  value_u + B_{v_t}\end{equation}
Where $W_{v_t} \in \mathbb{R}^{L_{W}\times d_{model}}$ and  $B_{v_t}  \in \mathbb{R}^{L_{B}\times d_{model}}$ is the tensor after embedding $value_t$, and $V_u \in \mathbb{R}^{L_{V}\times d_{model}}$ is the result after encoding $value_u$.

\begin{figure}[t]
	\centering
	\begin{minipage}{0.48\linewidth}
		\centering
		\includegraphics{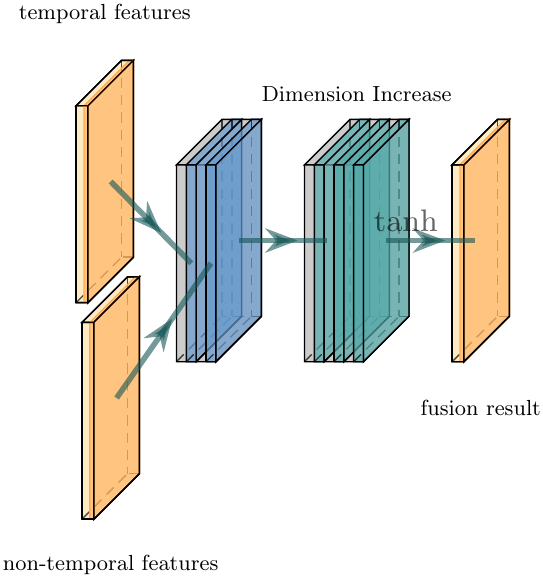}
		\caption{Temporal and non-temporal features fusion methods}
		\label{fig4}
	\end{minipage}
	\begin{minipage}{0.48\linewidth}
		\centering
		\includegraphics[scale=0.6]{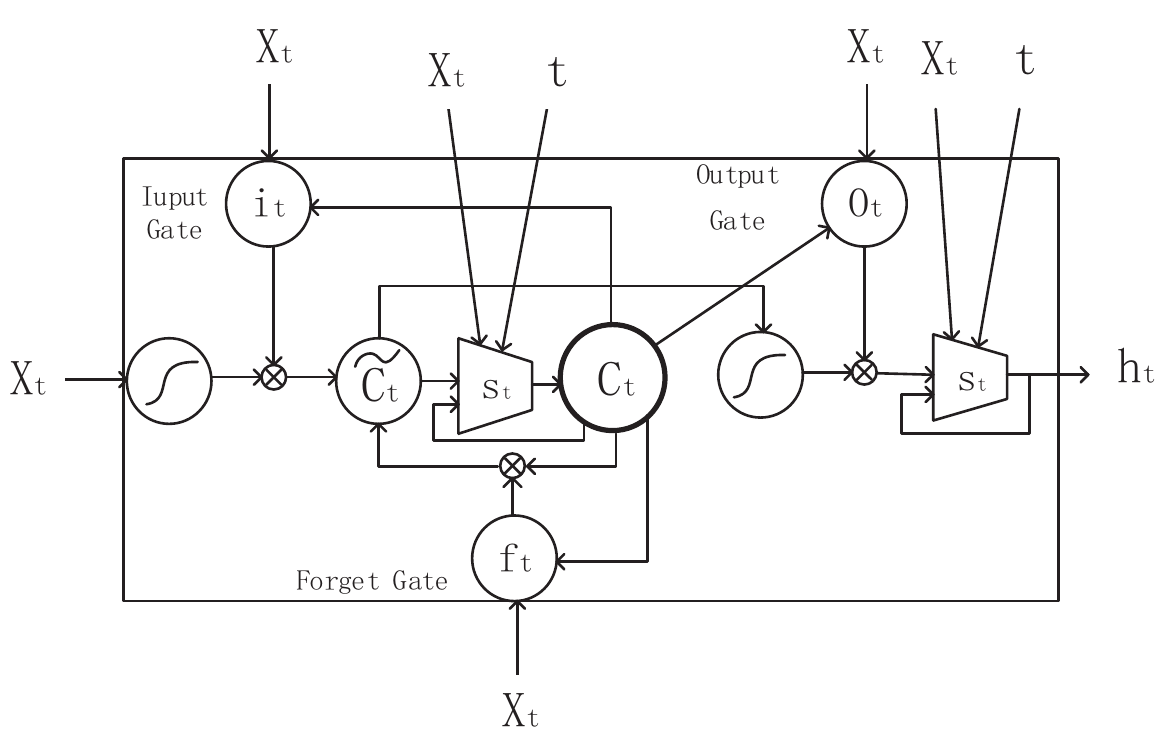}
		\caption{FG-LSTM Model}
		\label{fig3}
	\end{minipage}
\end{figure}

For the fusion of temporal and non-temporal features, many studies directly adopt the additive method, such as the most famous Transformer architecture \cite{vaswani2017attention}. The fusion of temporal and non-temporal features is not a simple additive relationship, so the method shown in the Fig.~\ref{fig4} is proposed. Firstly, stack the temporal and non-temporal features, then increase the dimension of the two features used $1 \times 1$ convolution structure. After processing by the $tanh$ activation function, we eventually fuse features into one-dimensional tensor on another $1 \times 1$ convolution structure.

Because the time interval between events is not equal, and the time of each event is a very important feature that can not be ignored. We add ``time encoding'' to the input embeddings and use two methods to encode time:

\subsubsection{Function Encoding}
We use sine and cosine functions of different frequencies just as ``positional encoding'' \cite{vaswani2017attention}:
\begin{equation}FE_{\left ( time, 2i \right )} = sin\left ( time / 10000^{2i/d_{model}}  \right )\label{eq1}
\end{equation}
\begin{equation}FE_{\left ( time, 2i+1 \right )} = cos\left ( time / 10000^{2i/d_{model}}  \right )\label{eq2}
\end{equation}
where $i$ is the dimension, $ i \in \left\{ 1, \ldots , d_{model}/2 \right\} $. That is, each dimension of the time encoding corresponds to a sinusoid. We chose this function because for any fixed time offset $k$, $FE_{time+k}$ can be represented as a linear function of $FE_{time}$. The time encoding have the same dimension $d_{model}$ as the embeddings, so that the two can be summed.

\subsubsection{Convolution Encoding}
We use the convolution structure to learn time encoding:
\begin{equation}\textbf{H}=tanh(Conv1d(T^L))\in \mathbb{R}^{L \times d_{model/2}}
\end{equation}
\begin{equation}\textbf{T}=tanh(Conv1d(\textbf{H}))\in \mathbb{R}^{L \times d_{model}}
\end{equation}
For the temporal features with length $L$ and dimension 1, that is, the size $L \times 1$, use the convolution kernel of $1 \times 1$ to learn the matrix with size $1 \times  {d_{model}}/{2}$, change the temporal features into the matrix with size $L \times {d_{model}}/{2}$. 
After the $tanh$ activation function, use the convolution kernel of $1 \times 1$ to learn the matrix with size ${d_{model}}/{2} \times d_{model}$ again, and change the temporal features into a matrix with the size of $L \times d_{model}$.
Finally get the temporal features with the size of $L \times d_{model}$ after $tanh$ activation function.

\subsection{Model Architecture}	

Long short-term memory (LSTM) \cite{hochreiter1997long} is an important ingredient for modern deep RNN architectures. The FG-LSTM extends the LSTM model by adding a new feature gate $s_t$, and the Fig.~\ref{fig3} shows the FG-LSTM model. The $x_t$ is the input features at time $t$, and others are basically consistent with ordinary LSTM. The feature gate has two factors: a feature filter and a time gate.

The combination of features and time gates only allows the features of certain kinds of features to be input into the neuron, and makes the neuron open only in a specific cycle. This ensures that each neuron will only capture the features of specific types of events and sample them, which solves the problem of poor training effect caused by the complexity and diversity of time and long event sequence.

The opening and closing of this feature gate is controlled by the features and time. Updates to the cell state $c_t$ and $h_t$ are permitted only when the gate is open. We proposed a particularly successful formulation of the feature gate as following:
\begin{equation}s_t=ReLU (W_{hs}\tanh (W_{xh}x_t+b_h)+b_s)\odot k_t \end{equation}
where $W_{xh}\in \mathbb{R}^{d_{model}\times h}, $$W_{hs}\in \mathbb{R}^{h\times s}$, $b_h\in \mathbb{R}^{1\times h}$ and $b_s\in \mathbb{R}^{1\times s}$ are the parameters to be learned, $h$ is hidden size, $s$ is output size. $ReLU$ and $tanh$ is the activation function, $x_t$ is the tensor input at time $t$, and $k_t$ is the time gate \cite{neil2016phased}.

Compared with traditional RNN and other excellent variants of RNN \cite{koutnik2014clockwork}, FG-LSTM can choose to update the learned parameters at the time point $t$ of irregular sampling. This allows the FG-LSTM to work with asynchronously sampled irregular time-series data. 
We can then rewrite the regular LSTM cell update equations for $c_t$ and $h_t$, using proposed cell updates $c_t^{\widetilde{}}$ and $h_t^{\widetilde{}}$ mediated by the feature gate $s_t$ :
\begin{align}
	&	i_{t}=\sigma _{i}(x_{t}W_{xi}+h_{t-1}W_{hi}+w_{ci}\odot c_{t-1}+b_{i}) \\
	&	f_{t}=\sigma _{f}(x_{t}W_{xf}+h_{t-1}W_{hf}+w_{cf}\odot c_{t-1}+b_{f}) \\	
	& c_t^{\widetilde{}}=f_t\odot c_{t-1}+i_t\odot \tanh_c(x_tW_{xc}+h_{t-1}W_{hc}+b_c)\\
	& c_t=s_t\odot c_t^{\widetilde{}}+(1-s_t)\odot c_{t-1}\\
	&	o_{t}=\sigma _{o}(x_{t}W_{xo}+h_{t-1}W_{ho}+w_{co}\odot c_{t}+b_{o}) \\
	& h_t^{\widetilde{}}=o_t\odot \tanh_h(c_t^{\widetilde{}})\\
	& h_t=s_t\odot h_t^{\widetilde{}}+(1-s_t)\odot h_{t-1}	
\end{align}
To sum up, for a neuron, only when it meets the type conditions of the corresponding feature gate, and the features information in its sampling period, neron will be updated. Therefore, it can be considered that this neuron represents the state of a certain type of features in a certain sampling period.
This is because the feature gate $s_t$, can be seen as a binary classifier to chose the cluster of features types responsible for each neuron. In addition, neurons do not update any information in the closing stage and maintain a perfect memory of past information, i.e. $c_j = c_{j-\Delta} $ if $k_t = 0$ for $t_{j-\Delta} \leq  t \leq  t_j$.
Therefore, other neurons that track other features can directly use the information of this set of features, even if they are far away from each other in sequence indexing. Because of this special mechanism, FG-LSTM can have much diverse and longer memory for modeling the dependency of multiple features.

We use a Softmax layer to predict the true label $y\widehat{}_t$ of the learned features tensor of sequence in the given decision times. This consists of two linear transformations with a $ReLU$ activation in the middle.
\begin{equation}y_t=softmax(max(0, h_tW_1+b_1)W_2 + b_2) \end{equation}
We use cross-entropy to calculate the classification loss of the prediction $y_t$ and true label $y\widehat{}_t$ of each sample as follows:
\begin{equation}Loss(y\widehat{}_t,y_t)=\frac{1}{L}\sum_{1\leq t\leq L}^{}(y^{\widehat{}}_t\times \ln y_t+(1-y^{\widehat{}}_t) \times \ln (1-y_t)) \end{equation}
We can sum up the losses of all the samples in one minibatch to get the total loss for back propagation.
\section{Experiments}
\begin{wraptable}{r}{8cm}
	\centering
	\caption{The Dataset Distribution}
	\label{tab:my-table}
	\begin{tabular}{|c|c|c|c|c|}
		\hline
		\textbf{Dataset  }      & \textbf{Target 0} & \textbf{Target 1} & \textbf{Total}  \\ \hline
		training set   & 475291   & 59328    & 534619 \\ \hline 
		validation set & 61698    & 6540     & 68238  \\ \hline
		evaluation set & 143373   & 19622    & 162995 \\ \hline
	\end{tabular}%
\end{wraptable}
The dataset used in this experiment is generated by Intensive Care Unit patient medical record data (MIMIC-III) of Beth Israel Deaconess Medical Center in the United States \cite{johnson2016mimic}. More than 20000 patient samples in MIMIC-III were extracted from the dataset, covering more than 4000 kinds and a total of more than 20 million multimodal irregular time-series data. In the experiment, the dataset is divided into training set,validation set and evaluation set, with a ratio of $7:1:2$.
 Table \ref{tab:my-table} shows the data distribution of the dataset, which is divided into two classes. All experiments were implemented by Pytorch \cite{paszke2019pytorch}, optimized by Adam optimization algorithm \cite{2014Adam}, with the learning rate of 0.0001 and the other parameters are selected as default parameters.  We set the random number seed to 1 to ensure the repeatability of the experimental results. Unless otherwise specified, $d_{model}$ (the dimension after features coding) is 256, the batchsize is 128. The detailed parameter settings of different experiments are described below. All the experiments are conducted on a single Nvidia RTX 3090 GPU (24GB memory), which is sufficient for all the baselines.

\subsection{Evaluating Metrics}
AUC (the area under Receiver Operating Characteristic curve) and AP (Average Precision) \cite{turpin2006user} are uesd in this paper. AUC is the area of ROC curve and the x-axis, and AP is the area of PRC (precision recall curve) and the x-axis, both of which are robust to the imbalanced data of positive and negative samples.

\subsection{Comparing Methods}
Because the proposed FG-LSTM is a variant based on the classical LSTM \cite{hochreiter1997long}, we choose the classical LSTM and three other excellent variants including BI-LSTM, Phase LSTM \cite{neil2016phased} and HE-LSTM \cite{LuchenLIU2019}. Recently, Transformer architecture has achieved the best performance in many problems, so we discuss the ability of Transformer related architecture to deal with multimodal irregular time-series. We chose the vanilla Transformer \cite{vaswani2017attention} and further select one of excellent variants in it called Informer \cite{zhou2021informer}. Because our experiment does not involve the generation process, therefore, only the encoder part of the Transformer architecture is used, and get the final output directly through a fully connected feed-forward network. For LSTM related architectures, only use one layer. For Informer, the number of layers in the original author's open source code is selected, that is, $n = 2$. For Transformer, in order to better compare with Informer, we selecte the same encoder layers as Informer. In addition, $d_{model}$ is changed to 256, which is consistent with LSTM architecture, and there is no change in the parameter settings of Transformer related architecture.

\subsection{Experimental Result}
\subsubsection {Non-temporal Features fusion methods}\label{section}

\begin{table*}[t]
	\centering
	\caption{Results of different non-temporal features fusion methods on different models, among them, different non-temporal feature fusion methods perform the best results, we use bold numbers in black, and underlined numbers are the best results in different models of the same fusion method.}
	\label{table2}
	\resizebox{\textwidth}{!}{%
		\begin{tabular}{|cc|c|c|c|c|c|c|c|c|}
			\hline
			\multicolumn{2}{|c|}{\textbf{Model}}                                        & \textbf{LSTM}  & \textbf{Bi-LSTM} & \textbf{Phased LSTM} & \textbf{HE-LSTM} & \textbf{Transformer} & \textbf{Informer} & \textbf{FG-LSTM}    & \textbf{Count}               \\ \hline
			\multicolumn{1}{|c|}{\multirow{2}{*}{\textbf{Our Method}}}   & \textbf{AUC} & \textbf{75.63} & \textbf{75.59}   & \textbf{72.52}       & 74.21            & \textbf{75.69}       & \textbf{76.05}    & \underline{\textbf{78.85}}      & \multirow{2}{*}{\textbf{12}} \\ \cline{2-9}
			\multicolumn{1}{|c|}{}                                       & \textbf{AP}  & \textbf{34.96} & \textbf{34.92}   & \textbf{30.45}       & 32.44            & \textbf{34.31}       & \textbf{34.93}    & \underline{\textbf{38.90} }     &                              \\ \hline
			\multicolumn{1}{|c|}{\multirow{2}{*}{\textbf{Other Method}}} & \textbf{AUC} & 68.59          & 70.36            & 68.95                & \textbf{76.35}   & 64.23                & 75.91             & \underline{76.37}               & \multirow{2}{*}{\textbf{2}}  \\ \cline{2-9}
			\multicolumn{1}{|c|}{}                                       & \textbf{AP}  & 25.94          & 26.85            & 26.63                & \textbf{34.80}   & 21.71                & 32.85             & \underline{36.42}               &                              \\ \hline
			\multicolumn{2}{|c|}{\textbf{Count}}                                        & 0              & 0                & 0                    & 0                & 0                    & 0                 & \textit{\textbf{4}} & -                            \\ \hline
		\end{tabular}%
	}
\end{table*}

In many previous studies, the processing methods of features from different feature spaces are only simple addition. According to this idea, a method is proposed as a comparative experiment, as shown below:
\begin{equation}x=V_e + sum\left(V_t \times V_u \right) \end{equation}
Where $V_e\in \mathbb{R}^{L_{e}\times d_{model}}$, $V_t\in \mathbb{R}^{L_{t}\times d_{model}}$ and $V_u\in \mathbb{R}^{L_{u}\times d_{model}}$ are the tensor encoded by $event$, $value_t$ and $value_u$ respectively.
In this experiment, the coding method without considering the temporary features. We uniformly choose the \eqref{eq1} \eqref{eq2} proposed above. For the fusion method of temporal features and non-temporal features, the addition method is directly selected, and the rest are discussed in detail below.

Table \ref{table2} shows the experimental results of AUC and AP on Table \ref{tab:my-table} dataset with different model architectures and different non-temporal features fusion methods.
It is obvious that, compared with the common methods, the proposed method of non-temporary features fusion has better performance. Except for the best performance in HE-LSTM framework, our proposed method has advantages in all other frameworks. The most obvious improvement is the Transformer framework, which has increased by $17.84\%$ in AUC and $58.03\%$ in AP. However, for the excellent Informer framework proposed for single-modal time-series, the improvement is not very obvious. The AUC and AP have only increased by $0.18\%$ and $6.33\%$ respectively, which shows that the Informer framework is not very sensitive to feature fusion methods. If we do not pay much attention to features fusion methods, Informer framework is indeed a good choice. For our proposed model FG-LSTM, the best performance of all models is obtained in different non-temporal feature fusion methods, and the AUC and AP are also improved by $3.24\%$ and $6.80\%$ respectively. Although the improvement is not very obvious, it also proves the superiority of our proposed model itself. In general, different feature fusion methods have great impact on the performance of different models, but excellent models are not particularly sensitive to feature fusion methods.

\subsubsection{Temporal and non-temporal features fusion methods}\label{section1}

The advanced of the proposed non-temporal features fusion method has been proved. Therefore, in this experiment, we verify the progressiveness of our proposed temporal and non-temporal features fusion method. In order to explore whether it is necessary to upgrade the dimension, we set up a group of control experiments to fuse the two features after stack directly with the help of $1 \times 1$ convolution kernel.
	\begin{table*}[t]
	\centering
	\caption{Results of different temporal features fusion methods on different models.}
	\label{tab3}
	\resizebox{\textwidth}{!}{%
		\begin{tabular}{|c|cccccc|cccccc|}
			\hline
			\multirow{3}{*}{\textbf{Model}} & \multicolumn{6}{c|}{\textbf{FE}}                                                                                                                                                                   & \multicolumn{6}{c|}{\textbf{CE}}                                                                                                                                                                             \\ \cline{2-13} 
			& \multicolumn{2}{c|}{\textbf{add}}                                    & \multicolumn{2}{c|}{\textbf{conv-add}}                                    & \multicolumn{2}{c|}{\textbf{$conv-add$}}          & \multicolumn{2}{c|}{\textbf{add}}                                         & \multicolumn{2}{c|}{\textbf{conv-add}}                                    & \multicolumn{2}{c|}{\textbf{$conv-add$}}               \\ \cline{2-13} 
			& \multicolumn{1}{c|}{\textbf{AUC}} & \multicolumn{1}{c|}{\textbf{AP}} & \multicolumn{1}{c|}{\textbf{AUC}}   & \multicolumn{1}{c|}{\textbf{AP}}    & \multicolumn{1}{c|}{\textbf{AUC}} & \textbf{AP} & \multicolumn{1}{c|}{\textbf{AUC}}   & \multicolumn{1}{c|}{\textbf{AP}}    & \multicolumn{1}{c|}{\textbf{AUC}}   & \multicolumn{1}{c|}{\textbf{AP}}    & \multicolumn{1}{c|}{\textbf{AUC}}   & \textbf{AP}    \\ \hline
			\textbf{LSTM}                   & \multicolumn{1}{c|}{75.63}        & \multicolumn{1}{c|}{34.96}       & \multicolumn{1}{c|}{\textit{76.88}} & \multicolumn{1}{c|}{\textit{36.38}} & \multicolumn{1}{c|}{76.87}        & 36.35       & \multicolumn{1}{c|}{\textbf{79.47}} & \multicolumn{1}{c|}{\textbf{39.50}} & \multicolumn{1}{c|}{79.09}          & \multicolumn{1}{c|}{38.56}          & \multicolumn{1}{c|}{79.18}          & 38.53          \\ \hline
			\textbf{Bi-LSTM}                & \multicolumn{1}{c|}{75.59}        & \multicolumn{1}{c|}{34.92}       & \multicolumn{1}{c|}{\textit{78.28}} & \multicolumn{1}{c|}{\textit{37.67}} & \multicolumn{1}{c|}{76.98}        & 37.46       & \multicolumn{1}{c|}{79.04}          & \multicolumn{1}{c|}{38.56}          & \multicolumn{1}{c|}{77.95}          & \multicolumn{1}{c|}{36.06}          & \multicolumn{1}{c|}{\textbf{79.64}} & \textbf{39.43} \\ \hline
			\textbf{Phased LSTM}            & \multicolumn{1}{c|}{72.52}        & \multicolumn{1}{c|}{30.45}       & \multicolumn{1}{c|}{\textit{74.82}} & \multicolumn{1}{c|}{\textit{33.57}} & \multicolumn{1}{c|}{72.83}        & 30.69       & \multicolumn{1}{c|}{76.34}          & \multicolumn{1}{c|}{34.03}          & \multicolumn{1}{c|}{74.54}          & \multicolumn{1}{c|}{32.90}          & \multicolumn{1}{c|}{\textbf{78.01}} & \textbf{36.89} \\ \hline
			\textbf{HE-LSTM}                & \multicolumn{1}{c|}{74.21}        & \multicolumn{1}{c|}{32.44}       & \multicolumn{1}{c|}{\textit{76.78}} & \multicolumn{1}{c|}{\textit{34.73}} & \multicolumn{1}{c|}{75.46}        & 33.24       & \multicolumn{1}{c|}{\textbf{77.06}} & \multicolumn{1}{c|}{\textbf{35.86}} & \multicolumn{1}{c|}{76.79}          & \multicolumn{1}{c|}{34.72}          & \multicolumn{1}{c|}{75.47}          & 34.37          \\ \hline
			\textbf{Transformer}            & \multicolumn{1}{c|}{75.69}        & \multicolumn{1}{c|}{34.31}       & \multicolumn{1}{c|}{\textit{76.14}} & \multicolumn{1}{c|}{\textit{34.88}} & \multicolumn{1}{c|}{75.85}        & 34.87       & \multicolumn{1}{c|}{\textbf{76.41}} & \multicolumn{1}{c|}{\textbf{35.19}} & \multicolumn{1}{c|}{75.65}          & \multicolumn{1}{c|}{34.22}          & \multicolumn{1}{c|}{76.26}          & 35.13          \\ \hline
			\textbf{Informer}               & \multicolumn{1}{c|}{76.05}        & \multicolumn{1}{c|}{34.93}       & \multicolumn{1}{c|}{\textit{76.19}} & \multicolumn{1}{c|}{\textit{35.66}} & \multicolumn{1}{c|}{75.96}        & 34.86       & \multicolumn{1}{c|}{\textbf{78.11}} & \multicolumn{1}{c|}{\textbf{35.29}} & \multicolumn{1}{c|}{75.99}          & \multicolumn{1}{c|}{34.90}          & \multicolumn{1}{c|}{76.05}          & 34.12          \\ \hline
			\textbf{FG-LSTM}                & \multicolumn{1}{c|}{\underline{78.85}}        & \multicolumn{1}{c|}{\underline{38.90}}       & \multicolumn{1}{c|}{\textit{\underline{80.67}}} & \multicolumn{1}{c|}{\textit{\underline{41.94}}} & \multicolumn{1}{c|}{\underline{78.47}}        & \underline{38.01}       & \multicolumn{1}{c|}{\underline{79.33}}          & \multicolumn{1}{c|}{\underline{39.09}}          & \multicolumn{1}{c|}{\textbf{\underline{81.20}}} & \multicolumn{1}{c|}{\textbf{\underline{42.69}}} & \multicolumn{1}{c|}{\underline{80.89}}          & \underline{41.59}          \\ \hline
			\textbf{Count}                  & \multicolumn{2}{c|}{0}                                               & \multicolumn{2}{c|}{0}                                                    & \multicolumn{2}{c|}{0}                          & \multicolumn{2}{c|}{8}                                                    & \multicolumn{2}{c|}{2}                                                    & \multicolumn{2}{c|}{4}                               \\ \hline
		\end{tabular}%
	}
\end{table*}

Table \ref{tab3} shows our experimental results. Where \textbf{FE} is function encoding, \textbf{CE} is convolution encoding, \textbf{add} is a direct addition method, \textbf{conv-add} is our method, and $conv-add$ is a comparative method without dimension upgrading. For the \textbf{FE} method without learning parameters, it can be seen that the \textbf{conv-add} method has achieved the most advanced experimental results in different models, while the $conv-add$ method without dimension upgrading is not as good as the \textbf{conv-add} method. But it is still better than the direct addition method in many models. For the \textbf{CE} method of learning parameters, it can be seen that no matter what kind of temporal and non-temporal feature fusion method, \textbf{CE} is better than \textbf{FE}, but none of the three feature fusion methods always has best performance in all models. Because our upgraded \textbf{conv-add} method also has parameters to learn, we believe that as long as the dataset is larger, the upgraded \textbf{conv-add} method can still be better than other methods in different models. Finally, for different models, different time coding methods and different feature fusion methods are used. Our FG-LSTM model is better than other models, which is enough to prove the robustness of our FG-LSTM. It also shows that the variants of LSTM are not necessarily inferior to the models of Transformer series.

\subsubsection{Experimental comparison of different length time-series}

\begin{figure}[t]
		\subfigure[The AUC of different models]{
	\begin{minipage}[t]{0.5\textwidth}
		\centering
		\includegraphics[scale=0.4]{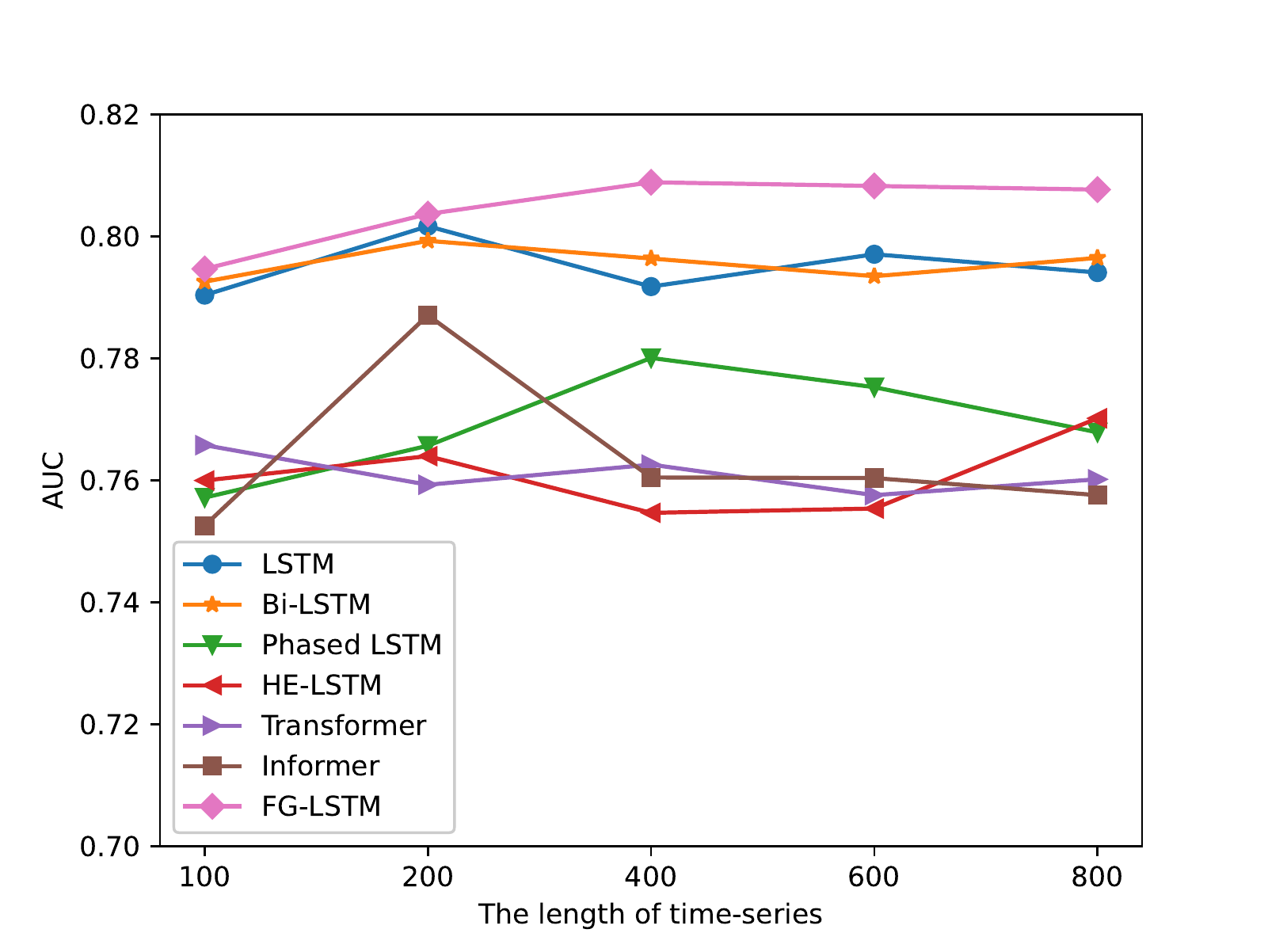}
		\label{figauc}
	\end{minipage}
}
	\subfigure[The AP of different models]{
	\begin{minipage}[t]{0.5\textwidth}
	\centering
	\includegraphics[scale=0.4]{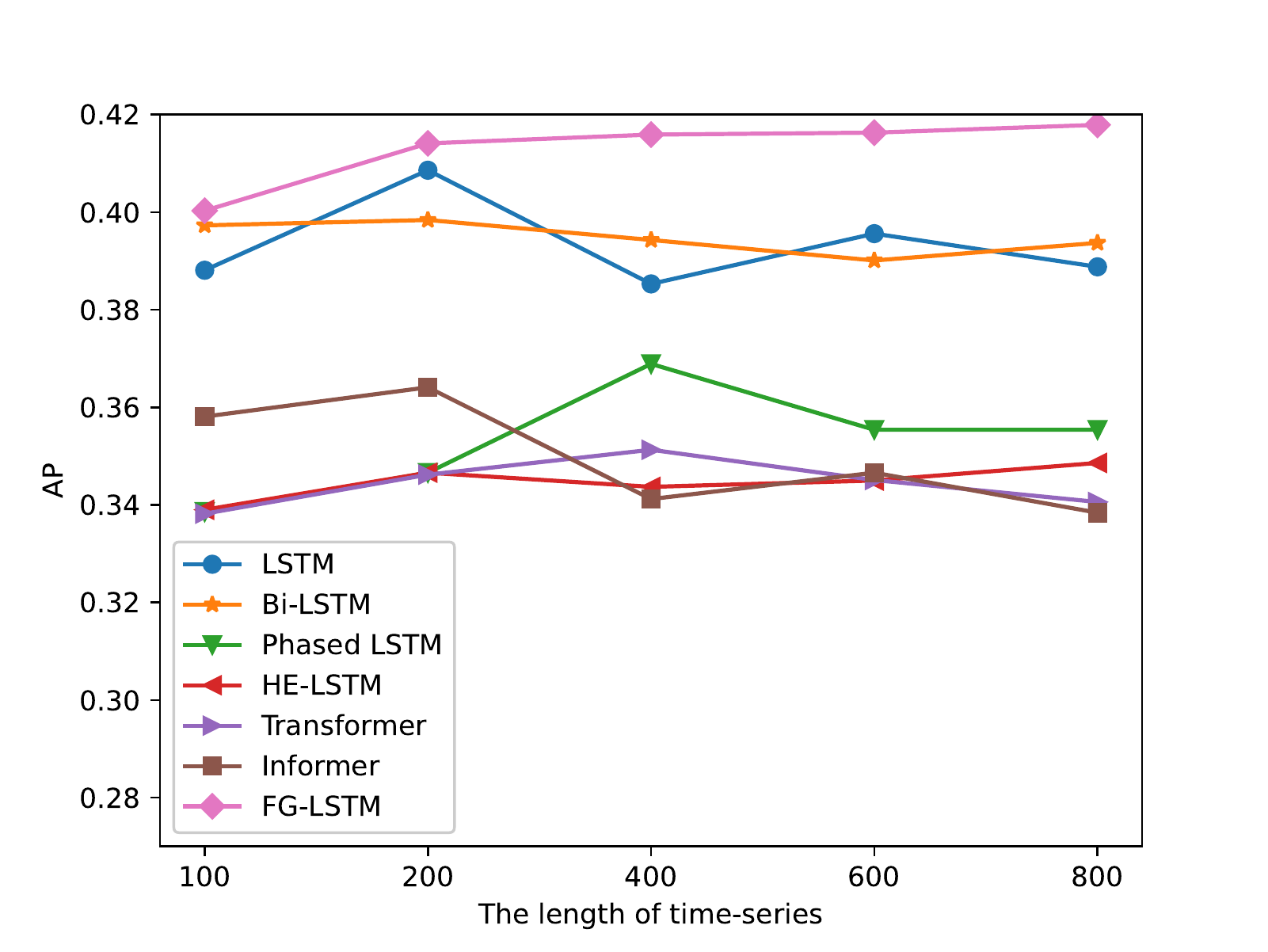}
	\label{figap}
	\end{minipage}
}
\caption{
The performance of different models on different time-series length.
}
\end{figure}

In order to verify the proposed model in this paper has stronger ability to capture the temporal dependence between features than other models, in this experiment, different models are input with different lengths of time-series data, ranging from 100 to 800. For the temporal feature coding method, choose \textbf{CE}. For the non-temporal feature fusion method, use our own method. For the temporal and non-temporal feature fusion method,  choose the $conv-add$ method without dimension upgrading. Fig.~\ref{figauc} and Fig.~\ref{figap} show the results of this experiment, for Transformer, when the length of time-series is 600 and 800, the Transformer failure for the out-of-memory, so we set the $batchsize$ is 64 to make 24GB memory enough. From the experimental results, we can draw the following conclusions:

Firstly, the time-series information is effective for the prediction results. When the input length is less than 400, most models will be improved with the increase of the length of the input sequence. Secondly, compared with other models, FG-LSTM is better at capturing the timing dependency in time-series. When the sequence length exceeds 400 and becomes longer and longer, the performance of the model is not improved much in AUC, but the AP is still improved steadily. However, other models can not capture the timing dependence under ultra long sequences, so they have not been greatly improved, and even the effect has become worse. Finally, we can see that the classical LSTM model is superior to Transformer and its variant model Informer, which shows that the time-series information extraction of Transformer series models is still slightly insufficient.

\section{Conclusion}
This paper proposes a features fusion framework and FG-LSTM model updated on the basis of LSTM. The model can well deal with multimodal irregular time-series data. At the same time, we also explore how to better encode time features and how to better integrate temporal features and non-temporal features, which is particularly important for irregular time-series data. Firstly, through the temporal features coding method and features fusion framework, the representation tensor obtained by the model can fuse the features and temporal dependency between different non-temporal information, effectively capture the temporal dependency under ultra long sequences and the feature information of a minority events. Then, input the representation tensor of the obtained time-series into the FG-LSTM, due to the existence of feature gates, the model can automatically adapt to the multi-scale sampling frequency of multi-source complex data, asynchronously track the temporal information and feature information of different events. Finally, the experiments demonstrate that the method proposed in this paper has better performance than other typical methods on real datasets. The method in this paper is promising to expand and popularize, and can be further migrated to diverse fields, especially for multi-source asynchronous sampling sensor data and behavior recording data.

%
%
%

\end{document}